\documentclass[sn-mathphys,Numbered]{sn-jnl}

\usepackage{graphicx}%
\usepackage{multirow}%
\usepackage{subfig}
\usepackage{amsmath,amssymb,amsfonts}%
\usepackage{amsthm}%
\usepackage{mathrsfs}%
\usepackage[title]{appendix}%
\usepackage{xcolor}%
\usepackage{textcomp}%
\usepackage{manyfoot}%
\usepackage{booktabs}%
\usepackage{algorithm}%
\usepackage{algorithmicx}%
\usepackage{algpseudocode}%
\usepackage{listings}%
\usepackage{natbib}
\setcitestyle{square, numbers}
\usepackage{ulem} 

\theoremstyle{thmstyleone}%
\theoremstyle{thmstyletwo}%

\theoremstyle{thmstylethree}%

\begin{document}

\title[An evaluation of CNN models and data augmentation techniques in hierarchical localization of mobile robots]{An evaluation of CNN models and data augmentation techniques in hierarchical localization of mobile robots}


\author*[1]{\fnm{Juan José} \sur{Cabrera}}\email{juan.cabreram@umh.es}

\author[1]{\fnm{Orlando José} \sur{Céspedes}}\email{orlando.cespedes@goumh.umh.es}

\author[1]{\fnm{Sergio} \sur{Cebollada}}\email{s.cebollada@umh.es}

\author[1,2]{\fnm{Oscar} \sur{Reinoso}}\email{o.reinoso@umh.es}

\author[1]{\fnm{Luis} \sur{Payá}}\email{lpaya@umh.es}

\affil[1]{\orgdiv{Institute for Engineering Research (I3E)}, \orgname{Miguel Hern\'andez University}, \orgaddress{\city{Elche}, \country{Spain}}}

\affil[2]{\orgdiv{Valencian Graduate School and Research Network for Artificial Intelligence (valgrAI)}, \orgaddress{\country{Spain}}}


\abstract{This work presents an evaluation of CNN models and data augmentation to carry out the hierarchical localization of a mobile robot by using omnidireccional images. In this sense, an ablation study of different state-of-the-art CNN models used as backbone is presented and a variety of data augmentation visual effects are proposed for addressing the visual localization of the robot. The proposed method is based on the adaption and re-training of a CNN with a dual purpose: (1) to perform a rough localization step in which the model is used to predict the room from which an image was captured, and (2) to address the fine localization step, which consists in retrieving the most similar image of the visual map among those contained in the previously predicted room by means of a pairwise comparison between descriptors obtained from an intermediate layer of the CNN. In this sense, we evaluate the impact of different state-of-the-art CNN models such as ConvNeXt for addressing the proposed localization. Finally, a variety of data augmentation visual effects are separately employed for training the model and their impact is assessed. The performance of the resulting CNNs is evaluated  under real operation conditions, including changes in the lighting conditions. Our code is publicly available on the project website \href{https://github.com/juanjo-cabrera/IndoorLocalizationSingleCNN.git}{https://github.com/juanjo-cabrera/IndoorLocalizationSingleCNN.git}.}




\maketitle

\section{Introduction}\label{introduction}

In the ever-evolving landscape of Artificial Intelligence (AI), Convolutional Neural Networks (CNNs) have become a fundamental pillar of the technology, with disruptive problem-solving capabilities. This kind of neural networks were originally conceived for image recognition tasks, but have quickly transcended their initial boundaries, establishing themselves as a versatile and powerful tool for tackling a wide range of challenges in a variety of domains \citep{lecun1995convolutional}.

\vspace{0.3 cm}
The increasing use of CNNs can be attributed to their high ability to recognise patterns from different sources of information. This ability has made them essential in a wide variety of applications, from image recognition \citep{krizhevsky2012alexnet, simonyan2014VGG} and object detection \citep{redmon2016yolo, ren2015faster} to semantic segmentation \citep{ronneberger2015UNET} and even natural language processing \citep{kim2014NLPl}. The success of such architectures is based on their ability to extract features from data, which allows solving high-level problems such as visual localization.

\vspace{0.3 cm}
 In this sense, some researchers have addressed visual localization by means of 360 degrees vision sensors due to its relatively low cost and the wide range of information they provide. When capturing images in real-world scenarios, especially in robotics applications, the environmental conditions can vary significantly. Consequently, addressing the visual localization could be particularly challenging due to different phenomena such as changes in illumination conditions. For this reason, understanding and addressing the effects of illumination changes are crucial for developing robust CNN models. 

 \vspace{0.3 cm}

 Related with the above information, the main objective of this work is to analyze the influence of different visual effects applied to the training data in order to carry out the mapping and localization of a mobile robot, which moves in an indoor environment under real operation conditions. For this purpose, the omnidirectional images captured by a catadioptric vision sensor are used to train a CNN. Both the raw images, and some sets of images obtained after introducing visual effects to the original images in a data augmentation process are considered during the training. In this paper, we have also evaluated the performance of state-of-the-art CNN models when addressing localization through a hierarchical approach. In this sense, the CNN will be adapted and re-trained with a dual purpose: (1) to perform a rough localization step in which the model is used to predict the room from which a test image was captured, and (2) to address the fine localization step, which consists in retrieving the most similar image of the visual map among those contained in the previously predicted room by means of a pairwise comparison between descriptors obtained from an intermediate layer of the CNN. The main contributions of this paper can be summarized as follows.

\begin{itemize}
\item A CNN is adapted and re-trained to predict the room from which the robot captured an omnidirectional image which is transformed into panoramic. This approach enhances robotic localization by initially performing room recognition.

\item We use the re-trained CNN to embed panoramic images into holistic descriptors by extracting the activation of an intermediate layer. These descriptors are compared to the visual model of the retrieved room via nearest neighbour search, providing an efficient method for scene recognition and position retrieval.

\item We conduct a thorough study of the individual influence of different data augmentation visual effects when training a model to perform hierarchical localization. This analysis contributes to improve the robustness of the model and its generalization ability in localization tasks.

\item We evaluate the performance of different state-of-the-art CNN models that are used as the backbone for the proposed localization task. This comparative evaluation provides valuable insights for selecting the most suitable CNN architecture for real-world localization applications.
\end{itemize}




This work is an extension of the initial developments presented in \cite{cespedes2023analysis}. In this previous work, we used a basic CNN model (Places \citep{zhou2014}) to perform the rough localization. However, our present proposal addresses both rough and fine localization steps and studies more exhaustively different state-of-the-art models such as AlexNet~\citep{krizhevsky2012alexnet}, ResNet-152~\citep{he2016resnet}, ResNeXt-101 64x4d~\citep{xie2017resnext}, MobileNetV3~\citep{howard2019mobilenetv3}, EfficientNetV2~\citep{tan2021efficientnetv2} and ConvNeXt Large~\citep{liu2022convnext}. Also, an ablation study of a variety of data augmentation visual effects are carried out with the aim of analysing the performance of the proposed tools under real operation conditions.

\vspace{0.3 cm}
The following sections are structured as follows. First, in the Section~\ref{sec:sota} we present a review of the state of the art on visual place-recognition and localization by means of artificial intelligence techniques. Second, in the Section \ref{sec:Method} we describe the proposed hierarchical localization method, the different CNN architectures which are evaluated and the proposed data augmentation visual effects. After that, we present in Section~\ref{sec:results} the dataset used and the experiments carried out to test and validate the proposed method. Finally, conclusions and future works are outlined in Section~\ref{sec:Conclusions}.

\section{State of the art}\label{sec:sota}

Artificial intelligence (AI) techniques are commonly proposed to address computer vision and robotics problems. Recent works, such as \citep{aguilar2017}, propose a pedestrian detector for Unmanned Aerial Vehicles (UAVs) based on Haar-LBP features combined with Adaboost and cascade classifiers with Meanshift. Another example is \citep{wang2018}, which utilizes an autoencoder for the fusion and extraction of multiple visual features from different sensors with the aim of carrying out motion planning based on deep reinforcement learning.

\vspace{0.3 cm}
CNNs have proven to be successful in many practical applications. Well-known architectures, such as GoogLeNet \citep{szegedy2015}, AlexNet \citep{krizhevsky2012alexnet} and VGG16 \citep{simonyan2014VGG} have been used as starting points to address new computer vision tasks. Regarding place-recognition, CNN models were firstly proposed to address this problem in \citep{chen2014convolutional}, where a pre-trained model called Overfeat \citep{sermanet2013overfeat} is used to extract features from images. \citet{sunderhauf2015performance} provided a thorough investigation on the performance of extracted features for place recognition. In fact, they found out that the features extracted from convolutional layers were more robust against different lighting conditions than those extracted from fully connected layers which outperformed towards viewpoint changes. \citet{bai2018cnn} propose the SeqCNNSLAM method, which consists in using the pre-trained AlexNet \citep{krizhevsky2012alexnet} to extract features and feed the SeqSLAM algorithm \citep{milford2012seqslam}. Also \citet{naseer2015robust} proposed a similar approach, but using GoogleNet \citep{szegedy2015}. Some of the works have not only used images as source of information, but also point clouds \citep{uy2018pointnetvlad} and both combined \citep{komorowski2021minkloc++}.

\vspace{0.3 cm}
In the context of robot localization, \citet{Kopitkov2018} propose using CNN holistic descriptors to estimate the robot position by learning a generative viewpoint-dependent model of CNN features with a spatially-varying Gaussian distribution. \citet{Sarlin2019} carry out a hierarchical modeling using a CNN, which extracts local features and holistic descriptors for 6-DOF localization. In that paper, a coarse localization is solved by using global descriptors, while a fine localization is solved by matching local features. Recent works \citep{cebollada2022development} have proposed hierarchical visual models for efficient localization. This method involves arranging visual information hierarchically in different layers so that localization can be solved in two main steps. The first step involves coarse localization to roughly determine the area where the robot is located, and the second step involves fine localization within this pre-selected area.

\vspace{0.3 cm}
Regarding the training of CNNs, a large and varied dataset is essential. Since a lack of a large enough datasets is quite common, Data Augmentation (DA) can be used to increase the training instances to avoid overfitting. As for the DA for a mobile robot localization task, it is essential to apply visual effects that may occur in real operation conditions to make the model robust against those effects. Considering as many effects as possible would increase the effectiveness of the CNN, but this would imply more processing power and memory. Numerous researchers have leveraged the data augmentation technique as a valuable tool to enhance the efficacy of their models. For example, \citet{Ding2016} train a CNN with three distinct types of data augmentation operations. Their investigation aims to enhance the performance of Synthetic Aperture Radar target recognition by achieving invariance against pose variations. Similarly, \citet{Salamon2017} present a CNN designed for environmental sound classification, accompanied by an audio data augmentation strategy. This augmentation approach is useful to mitigate the scarcity of data in this domain, contributing to improved model performance. Furthermore, \citet{perez2017} present a study about the effectiveness of data augmentation to solve the classification task. \citet{shorten2019} present a survey about the existing methods for data augmentation and related developments. Nonetheless, the previously proposed data augmentation methods do not exactly analyze the visual phenomena that can occur when the mobile robot moves through the target environment under real-operation conditions. Therefore, the present work performs a data augmentation analysis that focuses on a wide range of those specific visual effects.

\vspace{0.3 cm}
In light of the above information, the aim of this work is to analyze the influence of some visual effects to carry out data augmentation for CNN training to address a hierarchical localization \citep{cebollada2022development}. Hence, the efficiency of each visual effect will be assessed through the ability of the CNN model to robustly estimate the position where the image was captured. In addition, this work focuses on evaluating the performance of different well-known CNN models for both the coarse and fine localization steps. The first one consists in estimating the room where the image was taken by means of a classification final layer. The second one is addressed by extracting a global descriptor from an intermediate layer of the CNN and used to retrieve the most similar image that conforms the visual map. To address the proposed evaluation, the unique source of information is the set of images obtained by an omnidirectional vision sensor installed on the mobile robot, which moves in an indoor environment under real operation conditions.

\section{Methodology}\label{sec:Method}

\subsection{Hierarchical Localization Approach}

\begin{figure}[h]
    \centering
    \includegraphics[width=\textwidth]{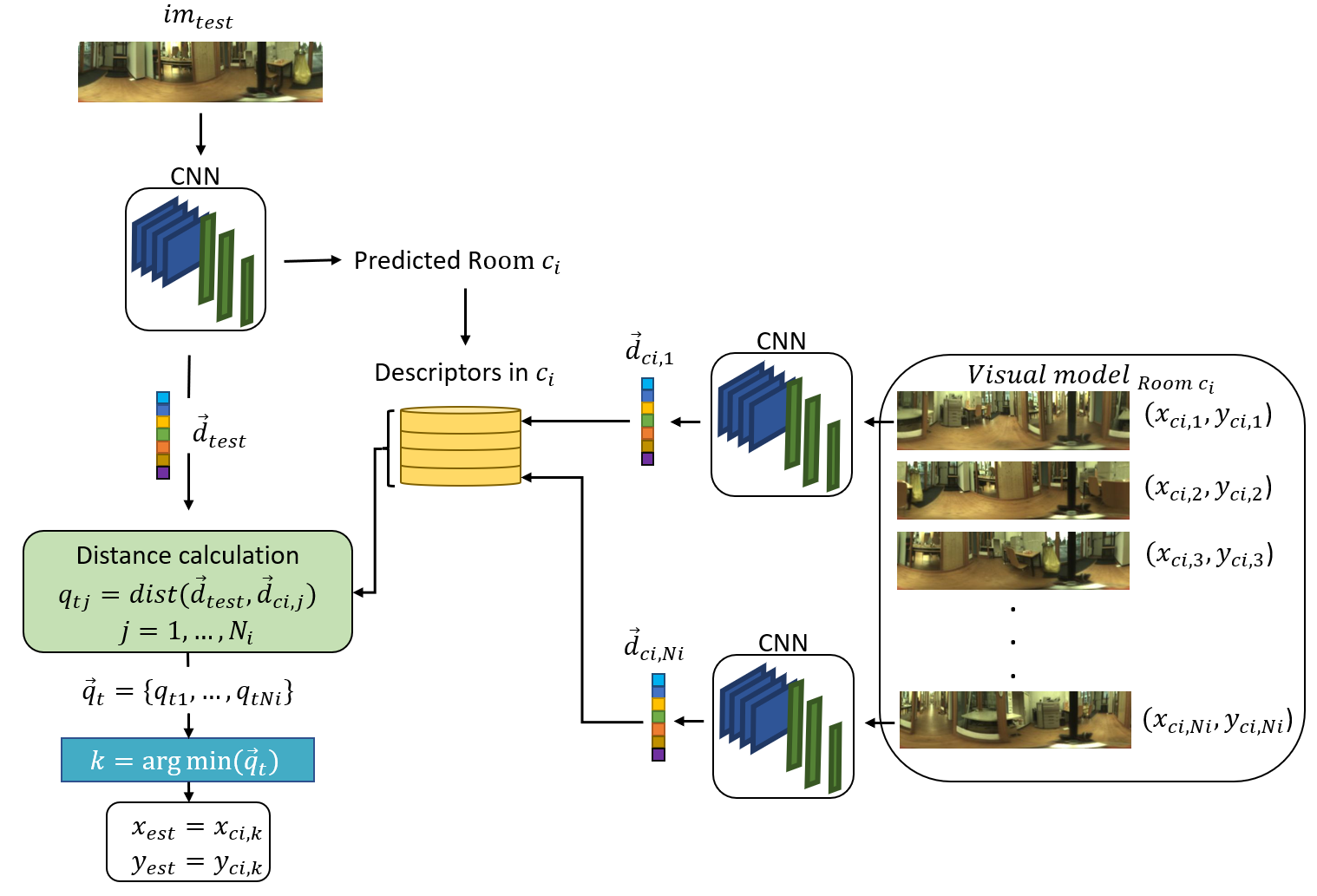}
    \caption{Diagram of the proposed hierarchical localization. The test image $im_{test}$ is the input of the CNN, which predicts the most likely room $c_i$ and embeds the image into a global descriptor $\vec{d}_{test}$ by flattening the last activation map. This descriptor is compared with the descriptors from the training dataset included in the retrieved room by means of a nearest neighbour search. Consequently, the capture point of the image that corresponds to the most similar descriptor ($im_{c_i,k}$) is considered an estimation of the position where $im_{test}$ was captured.}
    \label{fig:hierarchical_localization}
\end{figure}

This study aims to tackle visual localization through a hierarchical methodology by means of deep learning. The proposed approach (Figure \ref{fig:hierarchical_localization}) consists of two main steps: an initial stage for rough localization, which consist in identifying the room from which the test image has been captured, and a subsequent phase for fine localization where the position of the robot is obtained by a pairwise comparison between the test image and the visual model that conforms the pre-selected room.

\vspace{0.3 cm}
The initial step of rough localization is performed using the output of a CNN. The output layer of that CNN is composed by $R$ neurons, each one corresponding to a room ($R$ is the number of rooms or relevant areas in the target environment). Then, a SoftMax activation function is applied and the room prediction is obtained. However, before training the CNN, a dataset of labelled images captured along the target environment is needed. In this case, each image is labelled with the corresponding room information. The CNN is then trained to address the room retrieval task. Once the CNN is appropriately trained for the room classification task, the coarse localization step is performed: a test image $im_{test}$ is fed into the CNN and the output indicates the room $c_i$ in which the image was captured.

\vspace{0.3 cm}

Simultaneously, a holistic descriptor is extracted by flattening the activation map from the last convolutional layer. This descriptor $\vec{d}_{test}$ is compared with the descriptors $D_{c_i}=\{ \vec{d}_{c_i,1},\vec{d}_{c_i,2}, \ldots,\vec{d}_{c_i,N_{i}}\}$ from the visual map of the predicted room $c_i$, where $N_i$ is the number of images in the room  $c_i$. Note that the visual map descriptors are also obtained by flattening the last activation map of the same CNN. Then, the distance between the test descriptor $\vec{d}_{test}$ and each descriptor $\vec{d}_{c_i,j}$   $\epsilon$  $D_{c_i}$ in the room $c_i$ is calculated (Eq. \ref{eq:dist}).

\begin{equation}
\label{eq:dist}
    q_{t_j} = dist(\vec{d}_{test}, \vec{d}_{c_i,j}), \quad j = 1, \ldots, N_i
\end{equation}
where $N_i$ is the number of descriptors in room $c_i$ and $dist$ is the Euclidean distance (Eq. \ref{eq:euclidean})

\begin{equation}
\label{eq:euclidean}
    dist(\vec{d}_{test}, \vec{d}_{c_i,j}) = \sqrt{\sum_{i=1}^{m} (d_{test,i} - d_{c_i,j,i})^2}
\end{equation}

\noindent where $\vec{d}_{test} = (d_{test,1}, d_{test,2}, \ldots, d_{test,m})$ and $\vec{d}_{c_i,j} = (d_{c_i,j,1}, d_{c_i,j,2}, \ldots, d_{c_i,j,m})$ are the descriptors of size $m$, and $d_{test,i}$ and $d_{c_i,j,i}$ are the $i$-th components of the vectors $\vec{d}_{test}$ and $\vec{d}_{c_i,j}$, respectively.

\vspace{0.3 cm}
After that, a set $\vec{q}_t = \{ q_{t1}, \ldots, q_{tN_i} \}$ is constructed with the calculated distances. The index $k$ which minimizes the distance in the set $\vec{q}_t$ is found in Eq. \ref{eq:min}. Subsequently, the estimated position $(x_{est}, y_{est})$ corresponds to the position $(x_{c_i,k}, y_{c_i,k})$ from which the image $\vec{im}_{c_i,k}$ of the visual map (i.e, the image whose descriptor is the retrieved one $\vec{d}_{c_i,k}$) was captured (Eq. \ref{eq:coor}). This hierarchical approach ensures both a broad understanding of the scene and precise localization within the identified room, contributing to an effective visual localization strategy. Figure \ref{fig:hierarchical_localization} outlines the whole localization process.

\begin{equation}
\label{eq:min}
    k = \arg \min(\vec{q}_t)
\end{equation}

\begin{equation}
\label{eq:coor}
    x_{est} = x_{c_i,k}, \quad y_{est} = y_{c_i,k}
\end{equation}


\subsection{CNN selection and adaption}\label{backbone_description}
Designing a Convolutional Neural Network to a address a specific task supposes a big challenge. In the present work, the CNN must be able to predict the room in which an image was captured and embed the input image into a global descriptor to retrieve the exact position within the predicted room. Crafting a CNN from scratch demands both a profound understanding of the specificities involved and access to a sufficiently varied dataset for effective training. Furthermore, as previously demonstrated in \cite{ballesta2021}, in general terms, re-training networks that have been designed for a different objective yields more precise and reliable outcomes in the new task than training from scratch.

\vspace{0.3 cm}
In light of this information, this research work incorporates several widely recognised and tested CNN models, each of which serves as the backbone for our hierarchical localization task. These models cover a diverse range, addressing different architectural complexities and capabilities. All of the architectures employed were originally designed for visual object recognition. In this work, the CNN is first used to address the room retrieval problem, which is a similar task:

\begin{itemize}
    \item \textbf{AlexNet~\citep{krizhevsky2012alexnet}}: AlexNet is a pioneering CNN architecture known for its success in the ImageNet Large Scale Visual Recognition Challenge (ILSVRC) 2012. Comprising multiple convolutional and fully connected layers, AlexNet laid the foundation for subsequent CNN designs. This network and the following ones were trained to classify the 1.2 million high-resolution images into 1000 different classes. The weights and biases obtained by training with this database have been taken as starting point for our own task.
    \item \textbf{ResNet-152~\citep{he2016resnet}}: ResNet, or Residual Network, introduced the concept of residual learning. This approach is based on skip connections and allows the CNN to learn an identity function. ResNet-152 is a specific variant featuring 152 layers, enabling the model to effectively capture intricate hierarchical features. Although it is computationally costly due to its depth, its accuracy and robustness compensate this cost.
    \item \textbf{ResNeXt-101 64x4d~\citep{xie2017resnext}}: ResNeXt is an extension of the ResNet architecture, emphasizing a cardinality parameter to enhance model capacity. The cardinality is just the number of parallel blocks, that allows to learn various input representations. In this sense, ResNeXt-101 64x4d has a cardinality of 64. By increasing the cardinality, the network can capture a greater diversity of features, enhancing its potential ability to image recognition.
    \item \textbf{MobileNetV3~\citep{howard2019mobilenetv3}}: MobileNetV3 is designed for efficient mobile and edge computing applications. It uses depth-wise separable convolutions to build light weight deep neural networks. This fact makes them specially suitable for scenarios with resource constraints, such as performing the localization in real time by the robot's on-board computer.
    \item \textbf{EfficientNetV2~\citep{tan2021efficientnetv2}}: EfficientNetV2 is based on the EfficientNet architecture, and uses a technique called compound coefficient to scale up models in a simple but effective manner. It prioritizes model efficiency, achieving remarkable accuracy with fewer parameters compared to traditional CNNs. This makes EfficientNetV2 an attractive choice for applications requiring high accuracy with limited computational resources.
    \item \textbf{ConvNeXt Large~\citep{liu2022convnext}}: ConvNeXt Large represents a recent advancement in CNN architectures. It leverages a combination of depth-wise separable convolutions, an inverted bottleneck and spatial factorization ("patchify"), contributing to improved efficiency and effectiveness in capturing features. Thus, outperforming the previous models in terms of accuracy.
   
\end{itemize}

By evaluating these diverse CNN models, we aim to comprehensively understand their strengths and weaknesses in the context of scene recognition and localization task. Regarding the room recognition, the final layer of all the architectures needs to be adapted for classifying the images into $N$ categories corresponding to $N$ possible rooms in the target environment ($N$=9 in the dataset used in the present work, as described in Section \ref{sec:dataset}). As for the fine localization, the global descriptor has been extracted by flattening the output of the Average Pooling Layer of each CNN model. Finally, Table \ref{tab:models} shows a summary with the evaluated models and its corresponding number of FLOPs (Floating Point Operations) and the number of parameters.

\begin{table}[]
    \centering
    \caption{FLOPs and Parameters of the evaluated and adapted models when the size of the input image is 512x128x3 pixels.} 
    \label{tab:models}
    \begin{tabular}{@{}lcc@{}}
        \toprule
        \textbf{Backbone model} & \textbf{FLOPs} & \textbf{Number of Parameters} \\ \midrule
        AlexNet & 0.9 G & 57.0 M \\
        ResNet-152 & 15.2 G & 58.2 M \\        
        ResNeXt-101 64X4d & 20.4 G & 81.4 M \\        
        MobileNetV3 & 0.3 G & 4.2 M \\
        EfficientNetV2 & 16.2 G & 117.2 M \\
        ConvNeXt Large & 44.9 G & 196.2 M \\
        \bottomrule
    \end{tabular}
\end{table}

\subsection{ Data Augmentation}\label{data_augmentation}

Training a model involves setting up its parameters to perform a specific task. When a model has many parameters, it requires a sufficiently large number of examples for effective training. However, in practice, the training dataset is often limited. In such cases, data augmentation is a useful solution as it is able to generate new instances by applying various visual effects. This not only helps the model avoid overfitting but also makes it more robust against challenging real-operation dynamic conditions.

\vspace{0.3 cm}
In previous studies focused on training models for visual localization, various effects like changes in orientation, reflections, alterations in illumination, noise, and occlusions were applied \citep{cabrera2022}. The use of data augmentation has shown to improve model performance. These effects are applied individually or together to each image in the original dataset, and all the generated images are combined into a new augmented training dataset. However, the specific impact of each type of effect on the resulting CNN's performance is not well understood. This study aims to apply different data augmentation effects individually to evaluate their influence on the resulting CNN.

\vspace{0.3 cm}
The focus of this work is on two categories of visual effects: changes in illumination conditions and changes in orientation. For changes in illumination conditions, the following effects are considered:

\begin{itemize}
    \item \textbf{Spotlights and Shadows:} Circular light sources, like bulbs, are common indoors. The proposed approach involves increasing pixel values to simulate higher light intensity (spotlights) and decreasing pixel values to simulate shadows (shadow spots). Spotlights and shadow spots are applied separately for different data augmentation options. In our experiments, these bulbs are created with diameters ranging from 15 to 40 pixels. Five kinds of intensities variations are applied. In the first type the intensity is degraded +/- 160 and in the fifth +/- 100.

    \item \textbf{General Brightness and Darkness:} Low intensity values of the original images are increased to create brighter images, simulating higher overall illumination (e.g., a sunny day). Conversely, high intensity values are decreased to create darker images, simulating lower light supply (e.g., capturing images at night). Brightness and darkness are applied separately but used for the same data augmentation.
    
    \item \textbf{Contrast:} Image contrast plays a vital role in distinguishing objects in a scene. Images with low contrast tend to have a smoother appearance with fewer shadows and reflections. The contrast is modified following Eq. \ref{eq:contrast}
    \begin{equation}
        \label{eq:contrast}
        I_s=64+c*(I-64)
    \end{equation}

    
    where \(I_s\) is the resulting image,  \(I\) the original image and  \(c\) is the contrast factor. For  \(c>1\) the contrast increases and \(c<1\) decreases the contrast. 
    
    \item \textbf{Saturation:} Color saturation, indicating the color intensity given by pixels, is considered. Lower saturation results in less colorful images, potentially resembling grayscale images for very low saturation. This phenomenon may occur in real environments and is incorporated into data augmentation. The color saturation can be adjusted by first converting the RGB image to HSV. Then, the saturation channel can be directly modified by multiplying it by a constant factor \(s\). If the saturation is multiplied by \(s>1\), the colors become more saturated, whereas if multiplied by \(s<1\), the saturation decreases.
\end{itemize}

Regarding changes in orientation, these can occur during image capture when the robot captures images from the same position but with a different orientation. For this data augmentation option, new images are generated for each original image by applying rotations of $n$ degrees, where  \( n=i\times 10^{\circ}, i \in [1,35] \). Thus, for each original image in the training set, 35 additional images are generated.

\begin{figure}[h]%
    \centering
      \subfloat[]{%
       \includegraphics[width=0.47\linewidth]{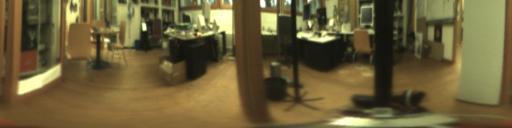}}
    \hfill
      \subfloat[]{%
       \includegraphics[width=0.47\linewidth]{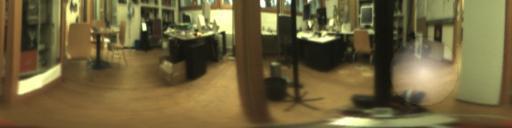}}
    \hfill
      \subfloat[]{%
       \includegraphics[width=0.47\linewidth]{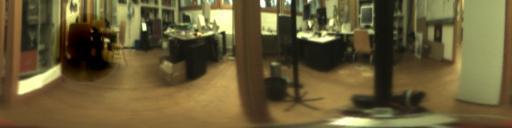}}
    \hfill
      \subfloat[]{%
       \includegraphics[width=0.47\linewidth]{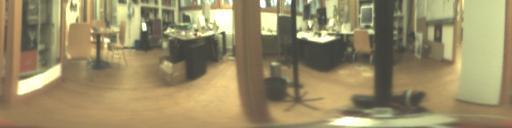}}
    \hfill
      \subfloat[]{%
       \includegraphics[width=0.47\linewidth]{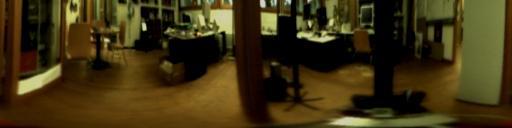}}
    \hfill
      \subfloat[]{%
       \includegraphics[width=0.47\linewidth]{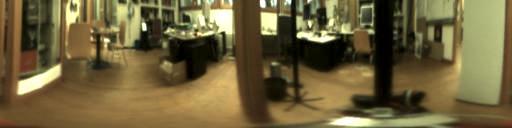}}
    \hfill
      \subfloat[]{%
       \includegraphics[width=0.47\linewidth]{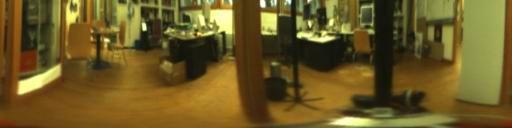}}
    \hfill
      \subfloat[]{%
       \includegraphics[width=0.47\linewidth]{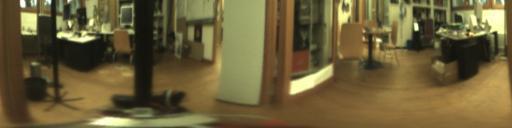}}
    \caption{Example of data augmentation where only one effect is applied over each image. (a) Original image, (b) spotlight effect, (c) shadow effect (d) general brightness, (e) general darkness, (f) contrast, (g) saturation and (h) rotation. The images contained in this dataset can be downloaded from the web site \href{https://www.cas.kth.se/COLD/}{https://www.cas.kth.se/COLD/}.}%
    \label{fig:data_augmenation_images}
\end{figure}

\vspace{0.3 cm}
Figure~\ref{fig:data_augmenation_images} shows an example of the effects applied to a sample omnidirectional image converted to panoramic format. The first image corresponds to the original one and the rest of images include the different effects presented above (they have been separately applied). 

\section{Results}\label{sec:results}

\subsection{COLD Freiburg Database}\label{sec:dataset}
The current study utilizes images sourced from the Freiburg dataset, a subset of the COLD (COsy Localization Database) database~\citep{pronobis2009}. This dataset contains omnidirectional images captured by a robot which follows various paths within a building at Freiburg University. The robot explores diverse spaces such as kitchens, corridors, printer areas, bathrooms, personal offices, and more. Image capture occurs under realistic operational conditions, including changes in furniture arrangement, the dynamic presence of individuals in scenes, and fluctuations in illumination conditions, including cloudy days, sunny days, and nights.

\vspace{0.3 cm}
To assess the impact of these variations on the localization task, we propose incorporating images taken exclusively on cloudy days as part of the training data. Additionally, a separate dataset comprising cloudy images (distinct from the aforementioned one) is employed as test set to evaluate localization performance without illumination changes. Furthermore, to appraise localization under varying illumination conditions, datasets captured on sunny days and at night are utilized as test sets. Beyond the images, the dataset offers ground truth data (obtained via a laser sensor), which is exclusively employed in this study to quantify localization errors. The ground truth over the path of the robot has been generated using the laser sensor in a grid-based SLAM technique, in particular, the one described in \citep{grisetti2005improving} and \citep{grisetti2007improved}. This solution, based on these two papers, can have an error up to 5 cm or 10 cm depending on the grid resolution.

\vspace{0.3 cm}
Concerning the image capture process, the robot acquires images while it moves, introducing potential blur effects or dynamic alterations. Moreover, the chosen environment has the longest trajectory within the available database and is characterized by extensive windows and glass walls, making visual localization a particularly challenging problem.  Consequently, this environment provides ideal conditions for evaluating the proposed localization methods under real operation conditions and real scenarios.

\vspace{0.3 cm}
The selected dataset contains images from nine distinct rooms: a kitchen, a bathroom, a printer area, a stairwell, a long corridor and four offices. The cloudy dataset is downsampled to achieve an average distance of 20 cm between consecutive image capture points, resulting in the Baseline Training Dataset comprising 556 images. This dataset serves the dual purpose of training the CNNs and providing a visual map. In addition, a Validation Dataset is used during training and keeps the same proportion of images as the Baseline Training set. The Validation Dataset is also sampled at 20-centimetre intervals, but in this case in an interleaved manner with respect to the Baseline Training Dataset in such a way that the images in the baseline and validation datasets are different. In this regard, the validation covers uniformly the whole environment, which is expected to be a robust approach for validation, considering that the retrained CNN must be able to solve the localization problem considering the whole environment. Furthermore, the Baseline Training Dataset undergoes a data augmentation as described in Section \ref{data_augmentation}, resulting in six additional training datasets. These datasets will be individually employed to train the CNNs, allowing an exploration of the impact of each visual effect on network performance. Table \ref{tab:train_room} shows a summary with the number of images per room of each training and validation dataset.

\vspace{0.3 cm}
In terms of the test data, various datasets are considered: Cloudy Test Dataset, comprising images captured in cloudy conditions along a route distinct from training and validation sets (2,595 images); Sunny Test Dataset, including all images captured in sunny conditions (2,114 images); and Night Test Dataset, containing all images captured at night (2,707 images). Table \ref{tab:test_room} shows a summary with the number of images per room of each test set. Consequently, network training and validation, in all instances, employs images captured exclusively in cloudy conditions, while testing occurs under three distinct lighting conditions: cloudy, sunny, or night. This methodology enables the assessment of the network's robustness against variations in lighting conditions.

\begin{table}[]
  \centering
  \caption{Number of images in each training dataset (number of images per room)}
  \label{tab:train_room}
  \begin{tabular}{@{}lccccccccc@{}}
    \toprule
    \textbf{\begin{tabular}[c]{@{}c@{}}Training\\ Dataset \end{tabular}} & \textbf{1P0-A} & \textbf{2P01-A} & \textbf{2P02-A} & \textbf{CR-A} & \textbf{KT-A} & \textbf{LO-A} & \textbf{PA-A} & \textbf{ST-A} & \textbf{TL-A} \\
    \midrule
    Baseline & 44 & 46 & 31 & 238 & 46 & 26 & 57 & 30 & 38 \\
    Validation & 43 & 47 & 32 & 236 & 46 & 26 & 57 & 31 & 38 \\
    Augmented 1 & 264 & 276 & 186 & 1428 & 276 & 156 & 342 & 180 & 228 \\
    Augmented 2 & 264 & 276 & 186 & 1428 & 276 & 156 & 342 & 180 & 228 \\
    Augmented 3 & 308 & 322 & 217 & 1666 & 322 & 182 & 399 & 210 & 266 \\
    Augmented 4 & 264 & 276 & 186 & 1428 & 276 & 156 & 342 & 180 & 228 \\
    Augmented 5 & 264 & 276 & 186 & 1428 & 276 & 156 & 342 & 180 & 228 \\
    Augmented 6 & 1364 & 1426 & 961 & 7378 & 1426 & 806 & 1767 & 930 & 1178 \\
    \bottomrule
  \end{tabular}
\end{table}

\begin{table}[]
  \centering
  \caption{Number of images in each test dataset (number of images per room)}
  \label{tab:test_room}
  \begin{tabular}{@{}lccccccccc@{}}
    \toprule
    \textbf{\begin{tabular}[c]{@{}c@{}}Test\\ Dataset \end{tabular}} & \textbf{1P0-A} & \textbf{2P01-A} & \textbf{2P02-A} & \textbf{CR-A} & \textbf{KT-A} & \textbf{LO-A} & \textbf{PA-A} & \textbf{ST-A} & \textbf{TL-A} \\
    \midrule
    Cloudy & 155 & 230 & 135 & 1040 & 254 & 177 & 222 & 133 & 249 \\
    Night & 168 & 215 & 168 & 1114 & 270 & 121 & 241 & 198 & 212 \\
    Sunny & 123 & 187 & 109 & 793 & 213 & 102 & 191 &  180 &  216 \\
    \bottomrule
  \end{tabular}
\end{table}

\subsection{Implementation details}
In this work, the CNNs are trained to address the coarse localization or room retrieval stage. As this is a classification task, these networks have been retrained employing a Cross Entropy loss function (Eq. \ref{eq:loss}).

\begin{equation}
\label{eq:loss}
    \mathcal{L}(y, \hat{y}) = -\frac{1}{B} \sum_{i=1}^{B} \sum_{j=1}^{R} y_{ij} \log(\hat{y}_{ij})
\end{equation}
\noindent where $y$ is the matrix of actual labels and $\hat{y}$ is the matrix of model predictions, both matrices have size $B \times R$, in which $B$ is the number of samples (batch size) and $R$ is the number of classes (rooms), $y_{ij}$ is 1 if sample $i$ belongs to class $j$ and 0 otherwise, and $\hat{y}_{ij}$ is the probability predicted by the model that sample $i$ belongs to class $j$.

\vspace{0.3 cm}
In addition, Stochastic Gradient Descent (SGD) with Momentum 0.9 and Learning Rate of 0.001 has been used as optimization algorithm. Furthermore, the training batch size ($B$) was 16 and the total number of epochs was 30. For every architecture, the network that presents the best validation accuracy for room retrieval during the training is preserved for testing. Table \ref{tab:training_parameters} summarizes all the values of the parameters that have been described above.

\vspace{0.3 cm}
All experiments are carried out with a NVIDIA GeForce RTX 3090 GPU with 24 GB. Our code is publicly available on the project website \href{https://github.com/juanjo-cabrera/IndoorLocalizationSingleCNN.git}{https://github.com/juanjo-cabrera/IndoorLocalizationSingleCNN.git}.

\begin{table}[]
  \centering
  \caption{Training Parameters for room retrieval}
  \label{tab:training_parameters}
  \begin{tabular}{@{}lc@{}}
    \toprule
    \textbf{Parameter}            & \textbf{Value}  \\ 
    \midrule
    Batch Size ($B$)                    & 16                  \\
    Number of Epochs              & 30                  \\
    Learning Rate         & $1 \times 10^{-3}$ \\
    Momentum & 0.9 \\
    Number of rooms ($R$) & 9 \\
    \bottomrule
  \end{tabular}
\end{table}

\subsection{CNN Backbone Ablation Study}
In this section, we asses an experimental evaluation of the different CNN models used as backbone presented in \ref{backbone_description} for both rough and fine localization.
As previously stated, the hierarchical localization proposed in this study comprises two distinct steps. The initial stage, rough localization step, involves retraining a model to execute the room retrieval task. Subsequently, the fine localization step utilizes the previously trained CNN to generate holistic descriptors, employing a nearest neighbor search method to estimate the precise position where an image was captured.

\subsubsection{Coarse Localization: Room retrieval} \label{sec:coarse_localization_backbones}

This section presents the results derived from the use of different CNNs for the execution of the coarse localization or room retrieval stage. As described in Section \ref{backbone_description}, the CNN models evaluated in this article are AlexNet~\citep{krizhevsky2012alexnet}, ResNet-152~\citep{he2016resnet}, ResNeXt-101 64x4d~\citep{xie2017resnext}, MobileNetV3~\citep{howard2019mobilenetv3}, EfficientNetV2~\citep{tan2021efficientnetv2} and ConvNeXt Large~\citep{liu2022convnext}. The reason why we have selected these models is to cover a wide range of architectures proposed for image classification in the last ten years.

\vspace{0.3 cm}


\begin{table}[h]
    \centering
    \caption{Room Retrieval ablation study for different top-level classification architectures tested under three different illumination conditions: cloudy, night, sunny and all together.}
    \label{tab:backbones_room_retrieval}
    \begin{tabular}{lcccc}
    \hline
    \multirow{2}{*}{\textbf{Backbone model}} & \multicolumn{4}{c}{\textbf{Room Retrieval Accuracy (\%)}}                                                                                        \\
                                    & \textbf{Cloudy} & \textbf{Night} & \textbf{Sunny} & \textbf{Global} \\ \hline
    AlexNet                         & 97.61                            & 97.60                           & 70.67                           & 89.93                            \\
    ResNet-152                      & 96.76                            & 96.64                           & 64.95                           & 87.63                            \\
    ResNeXt-101 64X4d               & 98.11                            & 95.16                           & 72.47                           & 89.71                            \\
    MobileNetV3                     & 98.50                            & 96.93                           & 77.29                           & 91.88                            \\
    EfficientNetV2                  & \textbf{98.81}  & 97.16                           & 75.73                           & 91.63                            \\
    ConvNeXt Large                  & 98.77                            & \textbf{97.64} & \textbf{86.28} & \textbf{94.80}  \\ \hline
    \end{tabular}
\end{table}

The results in Table \ref{tab:backbones_room_retrieval} showcase the performance of six different models used as backbone in the context of room retrieval across varied environmental conditions. In fact, each model was subjected to evaluation under cloudy, night, and sunny conditions, providing a comprehensive understanding of their robustness and adaptability to changes in environment illumination.

\vspace{0.3 cm}
AlexNet exhibits an excellent overall performance, particularly in Cloudy conditions with an accuracy of 97.61\%. In contrast, ResNet demonstrates robust performance but slightly lower accuracy compared to AlexNet. Notably, its accuracy decreases in sunny conditions which is the most demanding illumination environment. The ResNext model excels in cloudy conditions. However, it shows a comparatively lower accuracy in night scenarios. On the one hand, MobileNet stands out for its consistency, achieving high accuracy across all conditions. Its notable performance in sunny conditions, with an accuracy of 77.29\%, highlights its generalisation capability. On the other hand, EfficientNet emerges as a top-performing model, outperforming others in terms of accuracy in cloudy and night scenarios, which are the most similar to training conditions. Finally, the most striking result comes from ConvNext, which consistently achieves the highest accuracy in all scenarios, making it the top-performing model. Particularly noteworthy is its exceptional accuracy of 86.28\% in sunny conditions, indicating its robustness and generalization capabilities.

\subsubsection{Fine Localization}

\begin{figure}[h]
    \centering
    \includegraphics[width=\textwidth]{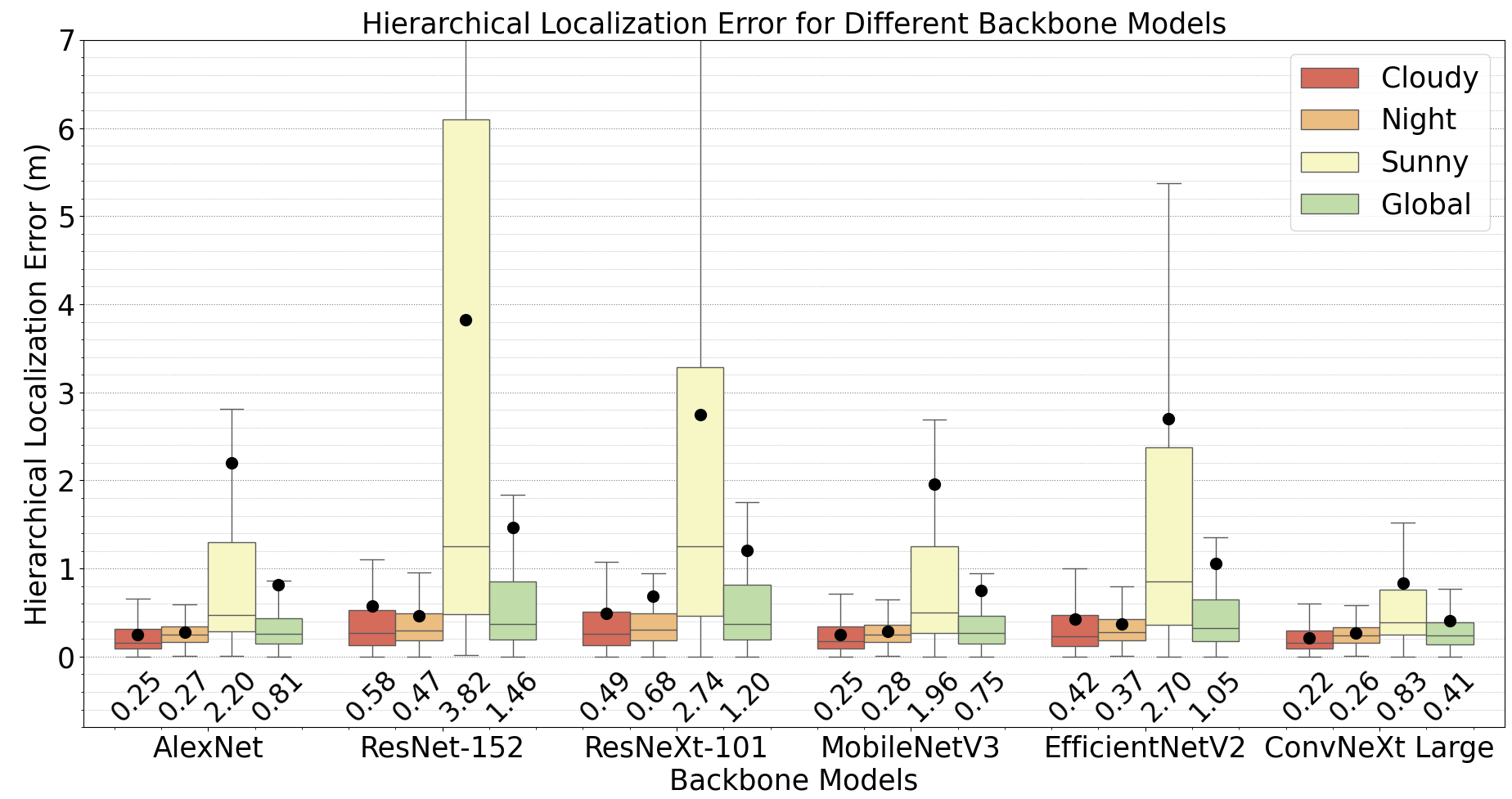}
    \caption{Hierarchical localization errors in meters for different CNN architectures. The box plots represent the distribution of errors, with whiskers indicating variability. The Mean Absolute Error for each model and condition is marked by a black dot and annotated with the specific error value. Results are obtained under different lighting conditions: cloudy (red), night (orange), sunny (yellow) and considering jointly the three conditions (green). }
    \label{fig:backbones:fine}
\end{figure}

Once the CNN model is trained for the room retrieval step, it can be used to embed the input image into a global descripor. This facilitates the resolution of the fine localization step through an image retrieval process, in which the descriptor of the test image is compared with the descriptors of the visual map of the previously retrieved room. As in previous subsection, we are going to evaluate the performance of different CNN backbones to address the fine lozalization step. Figure \ref{fig:backbones:fine} shows the hierarchical localization error for different backbone models (AlexNet, ResNet-152, ResNeXt-101, MobileNetV3, EfficientNetV2 and ConvNeXt Large) under various lighting conditions (cloudy, night, sunny) and considering jointly the three conditions (global). The errors are measured in meters and are represented by box plots with whiskers, indicating the distribution of the errors. Furthermore, the Mean Absolute Error (Eq. \ref{eq:mae}) is represented by the black dot and the text displaying the error value. In addition, Table \ref{tab:time_models} shows the computation time required to execute the whole hierarchical localization process for all the evaluated models.



\begin{equation}
\label{eq:mae}
\text{MAE} = \frac{1}{N} \sum_{i=1}^{N} \left| (x_i, y_i) - (\hat{x}_i, \hat{y}_i) \right|
\end{equation}

\noindent where $(x_i, y_i)$ is the actual position, $(\hat{x}_i, \hat{y}_i)$ is the position of the visual map retrieved after the complete localization process, and $N$ is the number of images in the test dataset.

\begin{table}[]
    \centering
    \caption{Computation Time required to execute the whole hierarchical localization process for all the evaluated models.}
    \label{tab:time_models}
    \begin{tabular}{@{}lc@{}}
        \toprule
        \textbf{Backbone model} & \textbf{Mean Time} \\ \midrule
        AlexNet & 3.4 ms\\
        ResNet-152 & 6.9 ms \\        
        ResNeXt-101 64X4d & 9.5 ms\\        
        MobileNetV3 & 4.6 ms \\
        EfficientNetV2 & 10.7 ms\\
        ConvNeXt Large & 12.5 ms \\
        \bottomrule
    \end{tabular}
\end{table}

\vspace{0.3 cm}
Each backbone model exhibited similar characteristics in hierarchical localization comparing to room retrieval, since both tasks are correlated. As Figure \ref{fig:backbones:fine} shows, AlexNet  demonstrated a consistent localization error and low dispersion for cloudy and night conditions. However, its performance degraded in sunny conditions. ResNet-152 displayed higher errors across all conditions compared to AlexNet, with a notable increase of both the mean absolute error and dispersion in sunny conditions. ResNeXt-101 demonstrated a better performance than ResNet-152 for cloudy and sunny conditions, but the error slightly increases for night scenarios. MobileNet consistently maintained low errors across all conditions, signifying its adaptability to diverse lighting environments. EfficientNet showcased a worse performance than MobileNet in each scenario. Finally, ConvNeXt emerged as the top-performing model, consistently outperforming others with the lowest errors across all conditions. Its remarkable accuracy in sunny conditions implies a robust capability to handle scenarios with substantial changes of the lighting conditions. In terms of computation time, Table \ref{tab:time_models} illustrates that the hierarchical localization process with the shortest average computation time occurs when employing AlexNet, which requires only 3.4 ms.  In contrast, the hierarchical localization process employing ConvNeXt Large requires the longest computation time, with a mean of 12.5 ms. However, despite the need for more time to estimate the position, this time is sufficiently short to enable real-time localization.

\subsection{Data Augmentation Ablation Study}

In this comprehensive experiment, the investigation is extended to evaluate the influence of both data augmentation effects (illumination and orientation changes) on the performance of the CNN. Due to the existence of a high probability of variations in robot orientation during operation under real operation conditions with respect to the images captured in the visual map, a model should demonstrate robustness to orientation changes. To this end, a data augmentation technique is employed that consists in applying 35 different orientation changes to each training image as described in Section \ref{data_augmentation}. This augmentation is essential to improve the adaptability of the model to the various orientations encountered in practice.

\vspace{0.3 cm}
Simultaneously, the illumination effects that occur under real operating conditions, a critical aspect for robust visual perception, have been explored in detail. Five specific lighting effects are considered (Section \ref{data_augmentation}): spotlights, shadow spots, general brightness/darkness, contrast, and saturation. Each effect is systematically applied individually on the training data set, leading to the creation of distinct augmented training datasets. Using the different effects separately allows a detailed understanding of their individual contributions, which sheds light on the importance of each effect in performance.

\vspace{0.3 cm}

In particular, for each image, the experiment incorporates a detailed approach by applying different levels of spotlights, contrast and saturation (five levels for each), ensuring a thorough assessment of the impact of these factors on the ability of the CNN to adapt to various lighting conditions. In addition, the effect of brightness is meticulously explored, with three levels of brightness and three levels of darkness applied to each image. This dual investigation of orientation changes and illumination effects is intended to provide a comprehensive understanding of the robustness of the CNN to cope with real-world challenges, encompassing variations in both spatial orientation and illumination conditions. As a result of applying these effects, six additional training datasets have been obtained: Augmented Training Dasaset 1 (spotlights), Augmented Training Dasaset 2 (shadows), Augmented Training Dasaset 3 (general brightness/darkness), Augmented Training Dasaset 4 (contrast), Augmented Training Dasaset 5 (saturation) and Augmented Training Dasaset 6 (rotations). Augmented Training Datasets 1, 2, 4 and 5 consist of 3336 images each, whereas Augmented Training Datasets 3 and 6 includes 3892 and 17236 images respectively. 

\vspace{0.3 cm}
In conclusion, in this ablation study the model is retrained using separately each of the Augmented Training Datasets 1, 2, 3, 4, 5 and 6. As in previous experiments, the Baseline Training Dataset serves as a visual map and the Validation Dataset is employed to validate the performance of the CNN. Furthermore, for the model evaluation, three different test datasets are considered: the Cloudy Test Dataset, the Night Test Dataset and the Sunny Test Dataset.

\subsubsection{Coarse Localization: Room retrieval}
In this subsection we use the best CNN architecture obtained in Section \ref{sec:coarse_localization_backbones}, which is ConvNeXt Large. In a similar approach, we have departed from the pre-trained weights for ImageNet Large Scale Visual Recognition Challenge and re-trained the model for the different datasets obtained by the proposed data augmentation.

\vspace{0.3 cm}
The Table \ref{tab:room_retrieval_accuracy_DA} presents the room retrieval accuracy when the model has been trained with each of the augmented training datasets previously described. The performance of the CNN is evaluated under the three different lighting conditions: cloudy, night, sunny and all together. 


\begin{table}[h]
    \centering
    \caption{Room retrieval accuracy for ConvNeXt Large architecture with different augmented training datasets.}

\begin{tabular}{lcccc}
\hline
\multirow{2}{*}{\textbf{Training Dataset}} & \multicolumn{4}{c}{\textbf{Room Retrieval Accuracy (\%)}}                               \\
                                           & \textbf{Cloudy} & \textbf{Night} & \textbf{Sunny} & \multicolumn{1}{l}{\textbf{Global}} \\ \hline
Baseline                                   & 98.77           & \textbf{97.64} & 86.28          & 94.80                               \\
Augmented 1 (Spotlights)                   & 98.84           & 97.45          & 86.14          & 94.71                               \\
Augmented 2 (Shadows)                      & 98.96           & 97.56          & 86.52          & 94.90                               \\
Augmented 3 (Brightness/Darkness)          & 98.81           & 97.41          & 91.11          & 96.10                               \\
Augmented 4 (Contrast)                     & 99.08           & 97.27          & \textbf{93.57} & \textbf{96.84}                      \\
Augmented 5 (Saturation)                   & 98.88           & 97.60          & 83.07          & 93.91                               \\
Augmented 6 (Rotations)                    & \textbf{99.15}  & 97.52          & 91.39          & 96.34                               \\ \hline
\end{tabular}

    \label{tab:room_retrieval_accuracy_DA}
\end{table}

Training with the baseline dataset shows a remarkable accuracy, especially in cloudy and night conditions. However, a significant decrease is observed in sunny conditions, which differ more from the training set. This evaluation provides a reference to analyse the impact of the different effects that have been applied to the training data. 

\vspace{0.3 cm}
The spotlight augmentation (Augmentation 1) shows insignificant improvements or even small decreases under night and sunny conditions. In contrast, data augmentation with shadows (Augmentation 2) produces slight improvements, especially in sunny conditions. 

\vspace{0.3 cm}
Alterations to the overall brightness and darkness of the image (Augmentation 3) are effective and show substantial improvements, especially in sunny conditions. In addition, contrast-based effects (Augmentation 4) are very effective, with substantial improvements in all lighting conditions and especially in sunny circumstances, thus achieving improved results in this challenging environment.

\vspace{0.3 cm}

Surprisingly, augmentation with changes in saturation (Augmented 5) shows a negative impact on accuracy, especially in sunny conditions. Finally, augmenting the data set with rotations (Augmented 6) shows substantial improvements, especially in cloudy conditions.

\subsubsection{Fine Localization}

Once the ConvNeXt Large model is trained for the room retrieval step, it can be used to embed the input image into a global descripor. This facilitates the resolution of the fine localization step through an image retrieval process, wherein the descriptor of the test image is compared with the descriptors of the visual map. As in previous subsection, we are going to evaluate the performance of different data augmentation effects to address the fine localization step.

\vspace{0.3 cm}

\begin{figure}[h]
    \centering
    \includegraphics[width=\textwidth]{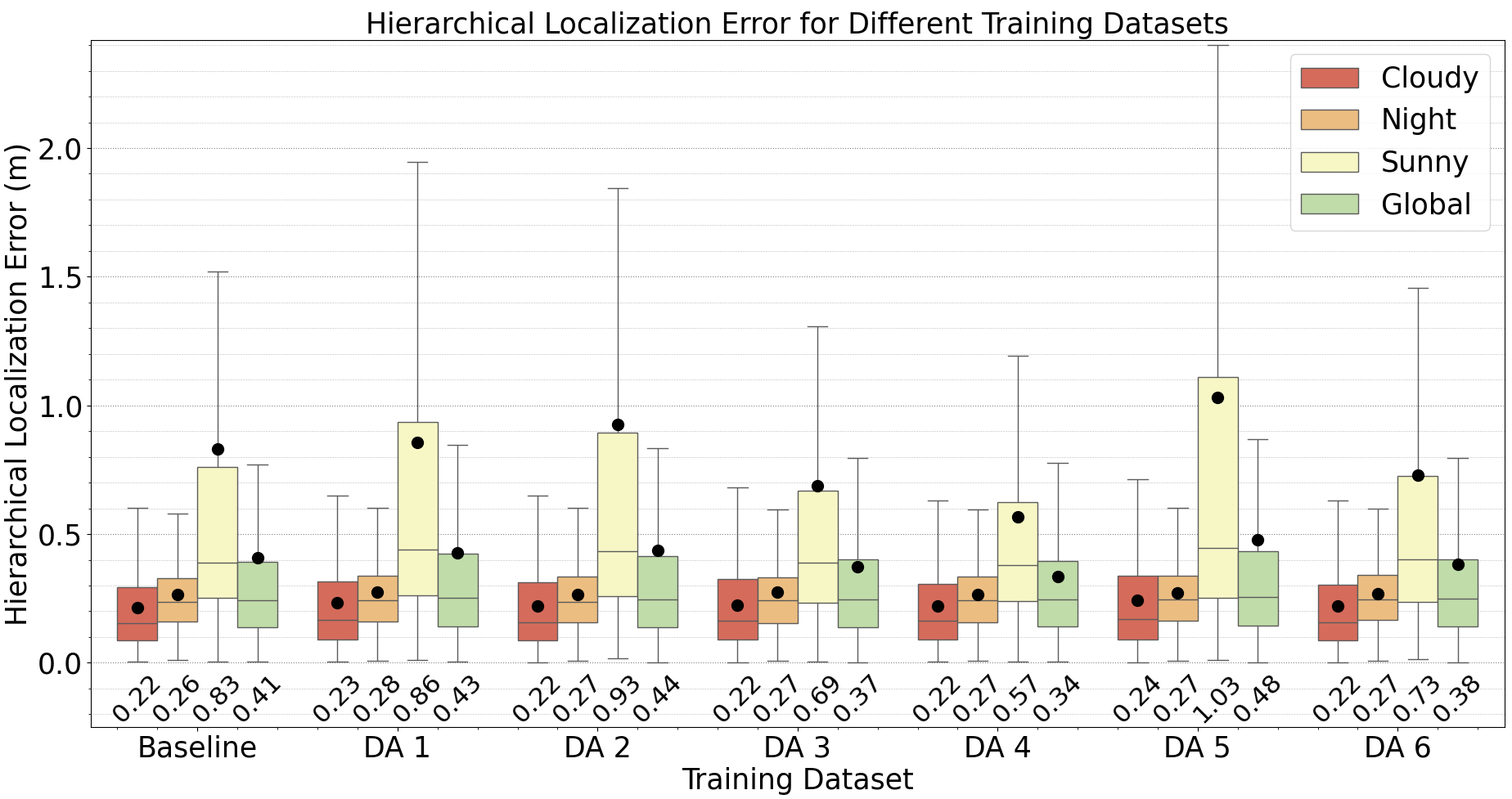}
    \caption{Hierarchical localization errors in meters when training the ConvNeXt Large architecture with different data augmentation effects. The box plots represent the distribution of errors, with whiskers indicating variability. The Mean Absolute Error for each model and condition is marked by a black dot and annotated with the specific error value. Results are obtained under different lighting conditions: cloudy (red), night (orange), sunny (yellow) and considering jointly the three conditions (green). }
    \label{fig:DA_fine}
\end{figure}

As shown in Figure \ref{fig:DA_fine}, training with every augmented dataset result in similar network performance under cloudy illumination conditions for the fine localization task, achieving a mean absolute error around 0,22 meters. The same happens under the night condition, in which the mean absolute error is around 0,27 meters. In this case, the minimum error is obtained by training the network without data augmentation.

\vspace{0.3 cm}
In contrast, under sunny lighting conditions the mean localization error has a higher variability, similarly to the coarse localization (Table \ref{tab:room_retrieval_accuracy_DA}). This demonstrates the correlation between the two tasks. Under this condition, the best fine localization result is obtained by training the model with the contrast effect (DA 4) and the worst with saturation (DA 5).

\subsubsection{General comparison with other methods}

Finally, the proposed method is compared with other previous global appearance techniques, including the use of single CNN structures (\citep{cabrera2022, rostkowska2023optimizing}), triplet structures \citep{alfaro2024hierarchical} and two classical analytical descriptors: HOG and gist, as described in \cite{cebollada2022development}. Both HOG and gist are only taken into consideration when testing with night and sunny conditions, since the conditions of the cloudy test experiment in \cite{cebollada2022development} are different to the conditions in the present work. Table \ref{tab:ComparacionMetodos} compares all the methods in a global localization task, using in all cases the COLD-Freiburg dataset, which is the same dataset used in the previous subsections. This table shows that ConvNeXt Large without data augmentation provides the best results in terms of localization error for cloudy and night conditions. Training with data augmentation does not improve the performance in cloudy conditions. However, it favours the results under sunny conditions. In this illumination condition, the best result is obtained with a triplet VGG16 proposed in \citep{alfaro2024hierarchical}. 

\begin{table}[!htb]
\centering
\caption{Comparison with other methods.}\label{tab:ComparacionMetodos}
\begin{tabular}{lccc}
\hline
\multicolumn{1}{c}{\textbf{\begin{tabular}[c]{@{}c@{}}Global-Appearance \\ Descriptor Technique\end{tabular}}} & \textbf{\begin{tabular}[c]{@{}c@{}}Cloudy\\ Error\end{tabular}} & \textbf{\begin{tabular}[c]{@{}c@{}}Night\\ Error\end{tabular}} & \textbf{\begin{tabular}[c]{@{}c@{}}Sunny\\ Error\end{tabular}} \\ \hline
Alexnet + DA \citep{cabrera2022}  & 0.29 m & 0.29 m  &  0.69 m\\ 
EfficientNet \citep{rostkowska2023optimizing}  & 0.24 m & 0.33 m  &  0.44 m\\ 
Triplet VGG16 \citep{alfaro2024hierarchical} & 0.25 m  & 0.28 m  & \textbf{0.40 m}  \\
ConvNeXt Large (ours) & \textbf{0.22 m}  & \textbf{0.26 m}  & 0.83 m  \\
ConvNeXt Large + DA (ours)  &  \textbf{0.22 m} & 0.27 m & 0.57 m \\ \hline
HOG \citep{cebollada2022development} & - & 0.45 m & 0.82 m \\
gist \citep{cebollada2022development} & - & 1.07 m & 0.88 m \\ \hline
\end{tabular}
\end{table}

\section{Conclusion}\label{sec:Conclusions}

This study assesses the application of a deep learning technique in addressing hierarchical localization using omnidirectional imaging. The technique involves training a CNN to perform room retrieval, addressed as an image classification problem. Additionally, the CNN is employed to embed the input image into a holistic descriptor from intermediate layers, aggregating  relevant information that characterizes the input image. Additionally, we evaluate the influence of two main components on the localization performance: CNN architecture and effects applied in the data augmentation.

\vspace{0.3 cm}
As for the CNN backbone, AlexNet shows excellent overall performance, especially when tested under the same lighting conditions than the training images. In contrast, ResNet performance decreases in sunny conditions which are the most challenging test conditions. This fact shows its low capability of generalization. The ResNext model surpass both in cloudy and sunny conditions, showcasing versatility across different lighting environments. However, EfficientNet exhibits a slight advantage over the ResNext model in terms of accuracy, although it requires more computational time. Furthermore, MobileNet consistently produces accurate results with a competitive computational time, demonstrating high performance across all conditions. Finally, the most striking result comes from ConvNext, which consistently achieves the highest accuracy in all scenarios, making it the top-performing model. Particularly noteworthy is its exceptional accuracy in sunny conditions, indicating its robustness and generalization capabilities.

\vspace{0.3cm}

Regarding the proposed data augmentation, training with the baseline dataset yields a remarkable accuracy, especially in cloudy and night conditions. However, a significant decrease is observed in sunny conditions, which diverge more from the training dataset. The spotlight effect shows marginal improvements, indicating that spotlight-based enhancement does not contribute to improve the generalization ability of the network. In contrast, data augmentation with shadows produces moderate improvements, especially in sunny conditions. Changing the overall brightness and darkness of the image produces substantial improvements, especially in sunny conditions. In addition, contrast-based effects are very effective, with significant improvements in all lighting conditions and especially in sunny conditions, improving results in this tough environment. Surprisingly, augmenting the dataset with changes in saturation shows a negative impact, especially in sunny conditions. Finally, increasing the dataset with rotations results in significant improvements in cloudy conditions. Finally, as for sunny conditions, the contrast effect yields the most optimal results, thereby enhancing the model's generalization capabilities and preventing overfitting. 

\vspace{0.3cm}
In future works, studying more advanced techniques for generating more realistic visual effects with Generative Adversarial Networks (GANs) is a priority. Furthermore, we will evaluate other deep learning schemas such as Siamese, Triplet Neural Networks and Feature Pyramid Networks (FPNs). Finally, we will approach the localization problem in outdoor environments by using CNNs, considering the specificities of such scenarios.

\backmatter
\bmhead{Code availability}
Our code is publicly available on the project website \href{https://github.com/juanjo-cabrera/IndoorLocalizationSingleCNN.git}{https://github.com/juanjo-cabrera/IndoorLocalizationSingleCNN.git}
\bmhead{Acknowledgments}

The Ministry of Science, Innovation and Universities (Spain) has supported this work through ``Ayudas para la Formación de Profesorado Universitario'' (FPU21/04969). This work is also part of the project TED2021-130901B-I00, funded by MCIN/AEI/10.13039501100011033 and the European Union ``NextGenerationEU”/PRTR, and of the project PROMETEO/2021/075 funded by Generalitat Valenciana.

\bibliography{sn-bibliography}

\end{document}